\documentclass[letterpaper]{article} % DO NOT CHANGE THIS
\usepackage{aaai24}  % DO NOT CHANGE THIS
\usepackage{times}  % DO NOT CHANGE THIS
\usepackage{helvet}  % DO NOT CHANGE THIS
\usepackage{courier}  % DO NOT CHANGE THIS
\usepackage[hyphens]{url}  % DO NOT CHANGE THIS
\usepackage{graphicx} % DO NOT CHANGE THIS
\usepackage{natbib}  % DO NOT CHANGE THIS AND DO NOT ADD ANY OPTIONS TO IT
\usepackage{caption} % DO NOT CHANGE THIS AND DO NOT ADD ANY OPTIONS TO IT
\usepackage{algorithm}
\usepackage{algorithmic}

\usepackage{newfloat}
\usepackage{listings}
\DeclareCaptionStyle{ruled}{labelfont=normalfont,labelsep=colon,strut=off} % DO NOT CHANGE THIS
\floatstyle{ruled}
\newfloat{listing}{tb}{lst}{}
\floatname{listing}{Listing}
\usepackage{inconsolata}
\usepackage{graphicx}
\usepackage{amssymb}
\usepackage{amsmath}
\usepackage{array}
\usepackage{caption}
\usepackage{subcaption}
\usepackage{enumitem}
\usepackage{booktabs}

\title{DenoSent: A Denoising Objective for Self-Supervised Sentence Representation Learning}
\author{
    Xinghao Wang,
    Junliang He,
    Pengyu Wang,
    Yunhua Zhou,
    Tianxiang Sun,
    Xipeng Qiu\footnote{Corresponding author}
}
\affiliations{
    School of Computer Science, Fudan University \\
    \{xinghaowang22, jlhe22, pywang22\}@m.fudan.edu.cn, \{zhouyh20, txsun19, xpqiu\}@fudan.edu.cn
}

\begin{document}

\frenchspacing

\maketitle

\begin{abstract}
Contrastive-learning-based methods have dominated sentence representation learning. These methods regularize the representation space by pulling similar sentence representations closer and pushing away the dissimilar ones and have been proven effective in various NLP tasks, e.g., semantic textual similarity (STS) tasks. However, it is challenging for these methods to learn fine-grained semantics as they only learn from the \textit{inter-sentence} perspective, i.e., their supervision signal comes from the relationship between data samples. In this work, we propose a novel denoising objective that inherits from another perspective, i.e., the \textit{intra-sentence} perspective. By introducing both discrete and continuous noise, we generate noisy sentences and then train our model to restore them to their original form. Our empirical evaluations demonstrate that this approach delivers competitive results on both semantic textual similarity (STS) and a wide range of transfer tasks, standing up well in comparison to contrastive-learning-based methods. Notably, the proposed intra-sentence denoising objective complements existing inter-sentence contrastive methodologies and can be integrated with them to further enhance performance. Our code is available at \url{https://github.com/xinghaow99/DenoSent}.
\end{abstract}

\section{Introduction}

Sentence representation learning is a fundamental task for natural language processing, which aims to embed sentence-level semantics into vectors of a fixed-sized $d$. High-quality sentence representations are expected to form a uniform space where similar semantics stay close, which is proven beneficial to various downstream tasks such as semantic textual similarity and information retrieval.

Transformer~\cite{vaswani2017attention}-based pre-trained language models (PLMs) like BERT~\cite{devlin2018bert} and RoBERTa~\cite{liu2019roberta} have shown remarkably superior performance on token-level tasks and can be adapted to various downstream tasks through fine-tuning, but they perform poorly in encoding sentence-level semantics due to the well-known anisotropy phenomenon in their representation space. Therefore, further training these PLMs for sentence-level representation learning remains a challenge.

Recently, contrastive methods have been adopted to sentence representation learning~\cite{gao2021simcse, yan2021consert, 2021DeCLUTR} and brought substantial improvement in both STS tasks and transfer tasks like sentiment analysis. These methods regularize the pre-trained language models (PLMs) representation space to be less anisotropic~\cite{ethayarajh2019contextual, wang2020understanding}, yielding competitive performance in downstream tasks.

However, one limitation of contrastive-learning-based methods is that their performance is highly dependent on the strategies of constructing positive pairs and selecting negative pairs. For instance, previous works adopted standard dropout~\cite{gao2021simcse}, different data augmentation strategies~\cite{yan2021consert} and different prompts~\cite{jiang2022promptbert} to construct positive pairs and may include a true-negative sample selection~\cite{zhou2022debiased} to alleviate the above problem. Nevertheless, contrastive methods solely learn the representation from the inter-sentence perspective, i.e., their supervision signal comes from the relationship between data samples, making it challenging to capture fine-grained semantics.

To address the above issue, we start from another perspective, i.e., the intra-sentence perspective, to learn sentence representations. In this work, we propose a novel denoising objective for sentence representation learning, which corresponds to another main branch of self-supervised learning~\cite{Liu_2021} other than contrastive, the generative branch, to provide intra-sentence supervision signals. Specifically, we adopt an encoder-decoder model structure that is identical to the original Transformer, except we only keep the encoded sentence representation to do cross-attention with a noisy version of the original sentence input. The training objective is to recover the noisy input to its original. Furthermore, the structure of our training framework has been designed to enable self-supervised integration of both intra-sentence and inter-sentence objectives.

Our main contributions can be summarized as follows:

\begin{enumerate}
    \item We propose a novel training objective to learn high-quality sentence representations from an intra-sentence perspective, i.e., utilize an auto-encoder structure and learn sentence representations by reconstructing the input sentence.
    \item We incorporate both discrete noises and continuous noises into our training framework, which facilitates our proposed denoising objective.
    \item We demonstrate that the proposed denoising objective is complementary to the contrastive objective, thereby proposing a promising sentence representation learning framework that combines both the intra-sentence and inter-sentence supervision signals.
\end{enumerate}

\section{Preliminaries}

\subsection{Sentence Representation Learning}

Sentence representations strive to encapsulate the underlying semantics and are adaptable for diverse applications. Each dense vector that represents a sentence enables direct measurement of semantic similarities, facilitates information retrieval, and supports training of classifiers tailored to diverse downstream tasks. There are two paradigms for generating sentence representations: frequency-based methods such as Bag-of-Words-based and TF-IDF-based and neural network-based methods like variants of Word2Vec~\cite{mikolov2013efficient,hill2016learning,kiros2015skip,logeswaran2018efficient} and variants of Transformer~\cite{2019Sentence,li2020sentence,su2021whitening,jiang2022promptbert}. Contrastive sentence representation learning~\cite{zhang-etal-2020-unsupervised,kim-etal-2021-self,meng2021cocolm,2021DeCLUTR,yan2021consert,gao2021simcse,janson2021semantic,zhou2022debiased,zhang2022contrastive} has become the main trend in this field for its effectiveness. On the other hand, generative methods of learning high-quality sentence representations~\cite{wang2021tsdae,wu2022sentence} are less investigated.

\subsection{Self-Supervised Learning}

Self-supervised learning is an ideal method for learning representations, owing to its intrinsic nature of not requiring any manual labels. It has been demonstrated to be effective across various modalities.~\cite{devlin2018bert, chen2020simple, schneider2019wav2vec}. There are principally two main branches of methods in self-supervised learning: \textbf{Contrastive learning} and \textbf{Generative learning}~\cite{Liu_2021, balestriero2023cookbook}.

Contrastive learning~\cite{sung2018learning} has been proven a promising approach in the field of sentence representation learning. The goal of contrastive learning is to pull semantically similar sentences closer together, while pushing dissimilar ones further apart in the representation space. For self-supervised contrastive learning, certain data augmentation strategies are necessary to form positive pairs, adhering to the principle of not using any labels. In the vision modality, methods such as cropping, resizing, rotation, and cutout are adopted to generate a positive sample from the input image. For contrastive sentence representation learning, ConSERT~\cite{yan2021consert} employs strategies such as adversarial attacks, token shuffling, cutoff, and dropout on the token embedding matrix to create positive samples. Meanwhile, SimCSE enhances performance by passing the same sentence into the pre-trained language model twice, thereby forming positive pairs. Contrastive learning has also been adopted as a pre-training objective for sentence representation learning.~\cite{wang2022text, su2022one}

Compared to contrastive learning, generative learning approaches are less investigated in the field of self-supervised sentence representation learning. Generative sentence representation learning attempts to generate original sentences from their corrupted or masked version ~\cite{yang2020universal, wang2021tsdae}. Recently, PaSeR\cite{wu2022sentence} was introduced, which auto-regressively generates important phrases from the original sentences; however, it necessitates the identification of these phrases beforehand.

\subsection{AutoEncoder}

AutoEncoder~\cite{kingma2013auto} is a neural network architecture that is designed to learn a compressed and efficient representation of the input data and it consists of two main components: an encoder and a decoder. The encoder maps the input data to a lower-dimensional representation, known as the bottleneck or latent representation. The decoder then reconstructs the bottleneck representation back to the original input space. Same to contrastive learning, autoencoders can also be trained in a self-supervised manner.

\begin{figure*}
\centering
\includegraphics[width=0.8\textwidth]{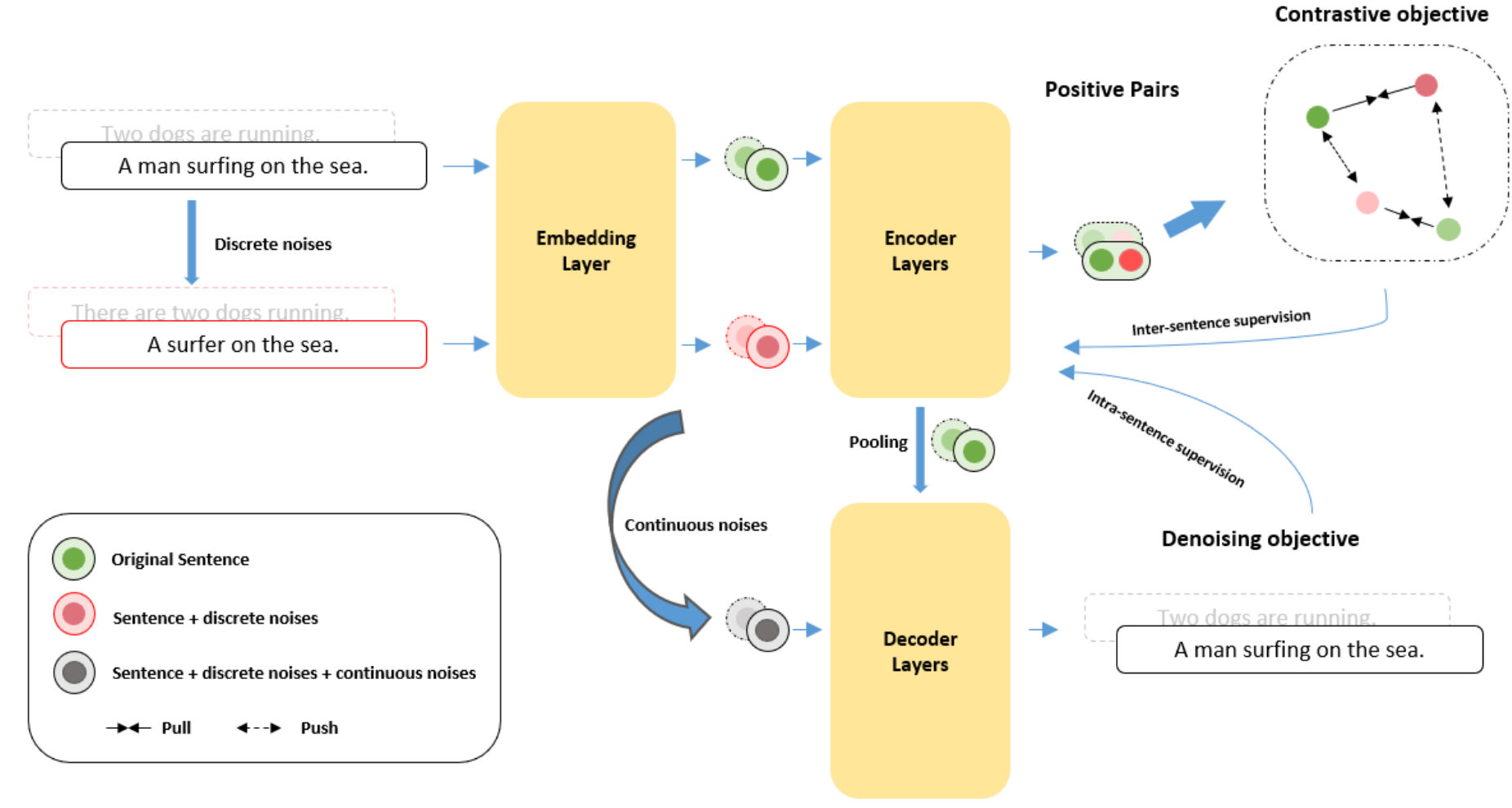}
\caption{Overview of DenoSent. The proposed sentence representation learning framework is a combination of two objectives, providing both inter-sentence and intra-sentence supervision signals. Note that we use pooling strategies to downsize the encoder outputs from [n\_tokens, hidden\_dim] to [1, hidden\_dim].}
\label{fig:overview}
\end{figure*}

\section{Methodology}

Figure. \ref{fig:overview} illustrates the overview of our proposed training framework, DenoSent. In this work, we aim to utilize intra-sentence supervision signals, using the original sentence as a guide. We achieve this by training an auto-encoder to reconstruct the original sentence from its noisy version. In our proposed training framework,  the auto-encoder closely mirrors the architecture of the original sequence-to-sequence Transformer. However, in our implementation, the length of the encoded source sequence is constrained to 1 through pooling (detailed in the implementation section), serving as the sentence representation. The decoder component is utilized exclusively during training and is subsequently discarded for evaluation. We introduce perturbations to the sentences in both the discrete and continuous space, and train our model to restore them to their original form from the perturbed sentences and their corresponding representations. We empirically demonstrate that our proposed denoising objective operates orthogonally to the contrastive objective, allowing both objectives to be seamlessly integrated into our framework. Experimental results reveal that the amalgamation of both intra-sentence and inter-sentence supervision signals yields competitive results on a broad range of tasks.

\subsection{Turn Vanilla Transformer into a Sentence Representation Learner}

The proposed denoising objective is both straightforward and efficacious. The following three modifications are made to the original Transformer~\cite{vaswani2017attention} model to turn it into a sentence representation learner:

\begin{itemize}
    \item Apply pooling strategies to reduce the length of the encoder output to 1, serving as the sentence representation, and seamlessly execute sequence-to-sequence learning.
    \item Discard the multi-head attention technique in the decoder and use single-head attention instead.
    \item In the prediction stage, use a denoising strategy to predict the original sentence rather than the standard auto-regressive technique.
\end{itemize}

As a sequence-to-sequence model, the vanilla Transformer first encodes an input sequence of symbol representations $\{x_1, ..., x_{n_1}\}$ to a sequence of continuous representations $z_x = \{z_1, ..., z_{n_1}\}$ through self-attention layers and feed-forward layers, where $n_1$ denotes the input sequence length. The Transformer decoder accepts a shifted-right target sequence of symbol representations $\{\langle s \rangle, y_1, ..., y_{n_2-1}\}$, where $\langle s \rangle$ denotes a start token for a sequence and $n_2$ for target sequence length, then transforms it to continuous representations $z_y$, and predict the target sequence $\{y_1, ..., y_{n_2}\}$. In the Transformer decoder, there is an additional attention layer other than the self-attention layer and the feed-forward layer in each block, which performs cross-attention operations across $z_x$ and $z_y$:

\begin{equation}
    CrossAttention(z_x, z_y)=softmax(\frac{z_y z_x^T}{\sqrt{d}})z_x
    \label{cross-attn-eq}
\end{equation}
Denote $d$ as the number of the hidden dimensions thus $z_x\in \mathbb{R}^{n_1 \times d}$, $z_y\in \mathbb{R}^{n_2 \times d}$ in the original Transformer. In the context of prediction, the Transformer model utilizes an auto-regressive approach to generate each token, where each generated token is dependent on preceding tokens:
\begin{equation}
    p(y)=\prod_{i=1}^n p(y_i|y_0,... ,y_{i-1})
    \label{ar-gen-eq}
\end{equation}
where $y_0$ denotes the start token $\langle s \rangle$.

In DenoSent, we employ pooling strategies on the encoder outputs to compress each sentence into a vector of a fixed-sized $d$. This can be alternatively viewed as reducing the input sequence length to 1, i.e., $z_x\in \mathbb{R}^{1 \times d}$ here.
After introducing certain perturbations to the input sentence, we feed the perturbed sentence into the Transformer decoder. Our model is then trained to reconstruct the original input sentence using solely the encoded sentence representation.
In the training process, $z_x$ obtains intra-sentence supervision signals in cross-attention operations(Eq. \ref{cross-attn-eq}) in the decoder and is forced to capture more semantic information to help recover the original sentence from its noisy version. Unlike the vanilla Transformer, which applies a causal mask on the attention matrix to facilitate auto-regressive training, DenoSent aims to predict each input sequence token based on the entire noisy sentence:

\begin{equation}
    p(x)=\prod_{i=1}^n p(x_i|\Tilde{x}_1,...,\Tilde{x}_n)
    \label{denoise-gen-eq}
\end{equation}
where $\Tilde{x}_i$ denotes the noisy version of token $x_i$.

Hence, let $\mathbf{S}$ be a corpus of sentences, the self-supervised denoising loss can be formed as:

\begin{figure*}[t]
\centering
\includegraphics[width=\textwidth]{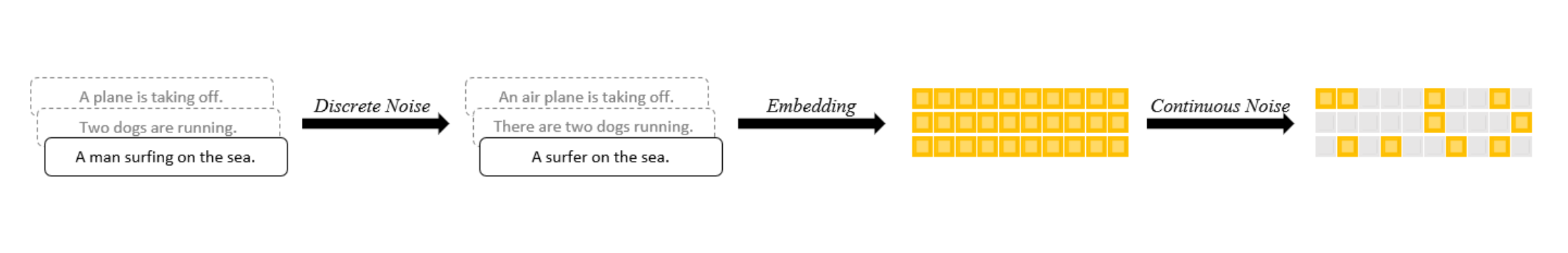}
\caption{The two-stage perturbation process wherein both discrete and continuous noises are sequentially incorporated into the original sentences. The discrete perturbation is achieved through back-translation or the use of a large language model (LLM), while the continuous perturbation is implemented by applying substantial dropout on the embedded sentences.}
\label{fig:noise}
\end{figure*}

\begin{equation}
    \ell_{denoising}=-\sum_{s_i}^S \sum_{j=1}^{s_i} log P(t_j|\Tilde{t}_1,...,\Tilde{t}_{s_i};\Theta)
    \label{denoise-objective}
\end{equation}
Here, $t_j$ represents the original token, while $\Tilde{t}_j$ denotes its noisy counterpart; $\Theta$ symbolizes the parameters of the model. The introduction of noise is detailed in the following subsection.

\subsection{The Perturbed Sentences: Discrete Noises and Continuous Noises}

Previous contrastive sentence representation learning techniques have employed a variety of data augmentation methods to construct positive pairs for contrastive learning. In the process, such operations introduce both discrete (e.g., token shuffling, token cutoff, \textit{inter alia}) and continuous (e.g., adversarial attack, dropout, \textit{inter alia}) noises to the original sentences, which enhance the generalization and alignment capabilities of the sentence encoder~\cite{yan2021consert, gao2021simcse}. In this work, we propose a two-stage perturbation strategy that integrates discrete noises and continuous noises sequentially(Figure. \ref{fig:noise}). These perturbations facilitate the generation of noisy input sentences, enabling us to train our sentence representation learner using the proposed denoising objective.

\subsubsection{Discrete Noises}

Discrete noises are introduced directly at the token level, resulting in a sequence of tokens $\{\Tilde{x}_1, ..., \Tilde{x}_{n'}\}$ derived from the original sequence $\{x_1, ..., x_{n}\}$. Simple token manipulations, such as deletion, swapping, or shuffling, have been shown to adversely affect performance, as they can disrupt the original semantics of the sentence.~\cite{yan2021consert, gao2021simcse} Here we propose to use two off-the-shelf data augmentation strategies to provide discrete noises, without compromising the inherent semantics of the original sentence. Specifically, we achieve this by leveraging the back-translation technique or a large language model (LLM) to rewrite the sentences. Machine translation aims to preserve original semantics in another language. By translating and back-translating sentences, we can obtain augmented sentences with similar semantics but varied syntax and expression. LLMs, on the other hand, can generate text based on the user's input and instructions after instruction fine-tuning~\cite{ouyang2022training, wei2022finetuned}. 
\citeauthor{cheng2023improving} have demonstrated that it is possible to generate sentence similarity labels for use in contrastive learning training, highlighting the ability of LLMs to capture sentence semantics. In our work, we exclusively use LLMs to rewrite the original sentences, introducing noise while preserving the underlying semantics. In practice, we utilize the pre-trained translation models for translation purposes, and OpenAI \texttt{gpt-3.5-turbo} for the instruction-following LLM. In our experiments, we discovered that employing the back-translation strategy results in marginally superior performance compared to using an LLM(See Table. \ref{tab:ablation_table}). Consequently, we adopt back-translation as the default strategy for incorporating discrete noises in the rest of the literature.

\subsubsection{Continuous Noises}

The introduction of continuous noises plays a crucial role in our proposed denoising objective, as it offers much greater control over the level of introduced noises within the continuous space. 
In our training framework, we employ dropout~\cite{srivastava2014dropout} at a substantial rate on the embedded sentences, setting most of the elements of the decoder input to zero. 
We subsequently train our model to reconstruct the sentence from the heavily corrupted input, drawing upon the output from the encoder, which serves as the sentence representation. This approach compels the model to retain sufficient semantic information in the encoded representation to facilitate the restoration of the original sentence.
The level of noise introduced can be controlled by the dropout rate, which determines the difficulty of the learning task.

\subsection{Combine with Contrastive Learning}
As the main trend in self-supervised sentence representation learning, contrastive learning~\cite{https://doi.org/10.48550/arxiv.2002.05709} has been proven effective in previous works~\cite{gao2021simcse}. The contrastive objective provides inter-sentence supervision signals by learning one sentence's representation from other sentences. Specifically, given a sentence $s$, a semantic-related positive example $s^+$ and a set of negative examples $s^-$ are needed to perform contrastive learning. Formally, denote $z$, $z^+$ and $z^-$ as the representation of $s$, $s^+$ and $s^-$, respectively, contrastive-learning-based methods utilize the InfoNCE~\cite{https://doi.org/10.48550/arxiv.1807.03748} loss:

\begin{equation}
\ell_{contrastive}=-log\frac{e^{sim(z, z^+)/\tau}}{\sum_{i=1}^N e^{sim(z, z_i^-)/\tau}}
\label{contrastive-objective}
\end{equation}
where $\tau$ denotes the temperature hyperparameter, $N$ is the number of negative samples for each training sample, and $sim$ for cosine similarity.

Unlike contrastive learning, the proposed denoising objective (as described in Eq. \ref{denoise-objective}) offers intra-sentence supervision signals by learning the representation directly from the sentence. Therefore, the denoising objective works independently from previous contrastive methods. Both objectives can be readily integrated:

\begin{equation}
    \ell = \ell_{contrastive} + \ell_{denoising}
    \label{combine-loss}
\end{equation}

For the contrastive objective, we add discrete perturbations to construct $s^+$ and in-batch negative samples $s^-$ for training. We reach our final results by optimizing Eq. \ref{combine-loss}.

\section{Experiment}

\begin{table*}[!ht]
\centering
% \fontsize{9pt}{10pt}\selectfont
\resizebox{\linewidth }{!}{
\begin{tabular}{ccccccccc}
\toprule
\textbf{Model} & \textbf{STS12} & \textbf{STS13} & \textbf{STS14} & \textbf{STS15} & \textbf{STS16} &\textbf{ STS-B} & \textbf{SICK-R} & \textbf{Avg.} \\
\hline
\hline
\multicolumn{9}{c}{\textit{Non-BERT Models}} \\
\hline
GloVe embeddings (avg.)$^\clubsuit$ & 55.14 & 70.66 &  59.73 & 68.25 & 63.66 & 58.02 & 53.76 & 61.32 \\
InferSent-GloVe$^\clubsuit$ & 52.86 & 66.75 & 62.15 & 72.77 & 66.87 & 68.03 & 65.65 & 65.01 \\
Universal Sentence Encoder$^\clubsuit$ & 64.49 & 67.80 & 64.61 & 76.83 & 73.18 & 74.92 & 76.69 & 71.22 \\
\hline
\hline
\multicolumn{9}{c}{\textit{BERT\&Post-Processing Models}} \\
\hline
BERT{\small base} (CLS)$^\blacksquare$ & 21.54 & 32.11 & 21.28 & 37.89 & 44.24 & 20.30 & 42.42 & 31.40 \\
BERT{\small base} (Mean)$^\blacksquare$ & 30.87 & 59.89 & 47.73 & 60.29 & 63.73 & 47.29 & 58.22 & 52.57 \\
BERT{\small base} (first-last avg.)$^\blacksquare$ & 39.70 & 59.38 & 49.67 & 66.03 & 66.19 & 53.87 & 62.06 & 56.70 \\
BERT{\small base}-flow$^\clubsuit$ & 58.40 & 67.10 & 60.85 & 75.16 & 71.22 & 68.66 & 64.47 & 66.55 \\
BERT{\small base}-whitening$^\clubsuit$ & 57.83 & 66.90 & 60.90 & 75.08 & 71.31 & 68.24 & 63.73 & 66.28 \\
\hline
\hline
\multicolumn{9}{c}{\textit{Contrastive-based Models}} \\
\hline
ConSERT-BERT{\small base}$^\heartsuit$ & 64.64 & 78.49 & 69.07 & 79.72 & 75.95 & 73.97 & 67.31 & 72.74\\
SimCSE-BERT{\small base}$^\clubsuit$ & 68.40 & 82.41 & 74.38 & 80.91 & 78.56 & 76.85 & 72.23 & 76.25 \\
DCLR-BERT{\small base}$^\blacksquare$ & 70.81 & 83.73 & 75.11 & 82.56 & 78.44 & 78.31 & 71.59 & 77.22 \\
DiffCSE-BERT{\small base}$^\diamondsuit$ & 72.28 & 84.43 & 76.47 & 83.9 & \textbf{80.54} & 80.59 & 71.23 & 78.49 \\
PromptBERT-BERT{\small base}$^\heartsuit$ & 71.56 & \underline{84.58} & 76.98 & \textbf{84.47} & 80.6 & \textbf{81.6} & 69.87 & 78.54 \\
SNCSE-BERT{\small base}$^\spadesuit$ & 70.67 & \textbf{84.79} & \underline{76.99} & 83.69 & \underline{80.51} & \underline{81.35} & \textbf{74.77} & \underline{78.97} \\
DenoSent-BERT{\small base}(contrastive only) & \underline{73.09} & 82.19 & 75.56 & 83.51 & 79.38 & 80.10 & 71.86 & 77.96\\
\hline
\hline
\multicolumn{9}{c}{\textit{Generative-based Models}} \\
\hline
CMLM-BERT{\small base}$^\blacklozenge$ & 58.20 & 61.07 & 61.67 & 73.32 & 74.88 & 76.60 & 64.80 & 67.22 \\
PaSeR-BERT{\small base}$^\blacklozenge$ & 70.21 & 83.88 & 73.06 & 83.87 & 77.60 & 79.19 & 65.31 & 76.16 \\
DenoSent-BERT{\small base}(generative only) & 69.50 & 83.83 & 75.09 & 82.78 & 77.75 & 77.59 & 66.78 & 76.19 \\
\hline
\hline
\multicolumn{9}{c}{\textit{Generative+Contrastive Models}} \\
\hline
\textbf{DenoSent-BERT{\small base}} & \textbf{75.57} & 83.77 & \textbf{77.24} & \underline{84.30} & 79.51 & 80.81 & \underline{74.09} & \textbf{79.33}\\

\bottomrule
\end{tabular}
}
\caption{Evaluation performance on 7 STS tasks. The reported metric is spearman correlation($\times 100$) based on cosine similarity following previous works. Bolded results and underlined results correspond to the best and second-best results in the same dataset, respectively. $\clubsuit$: results from ~\citealp{gao2021simcse}. $\blacksquare$: results from ~\citealp{zhou2022debiased}. $\diamondsuit$: results from \citealp{chuang2022diffcse}. $\heartsuit$: results from ~\citealp{jiang2022promptbert}. $\spadesuit$: results from \citealp{wang2022sncse}. $\blacklozenge$: results from ~\citealp{wu2022sentence}.}
\label{sts-exp}
\end{table*}

In our study, we evaluate the effectiveness of DenoSent on a variety of sentence-level tasks, including semantic textual similarity (STS), reranking, retrieval, and classification. To assess performance on STS tasks, we employed the SentEval toolkit~\cite{conneau2018senteval}, in line with previous research. The remaining tasks were evaluated using the Massive Text Embedding Benchmark (MTEB) toolkit~\cite{muennighoff2022mteb}.

\subsection{Datasets}

\textbf{Semantic textual similarity tasks.} STS tasks assess sentence similarity. Given a sentence pair, the similarity score is calculated based on the model-generated sentence representations, which is then compared to human-annotated similarity. We evaluate DenoSent on 7 STS tasks: \textbf{STS 2012–2016}(\citealp{agirre-etal-2012-semeval}, \citeyear{agirre-etal-2013-sem}, \citeyear{agirre-etal-2014-semeval}, \citeyear{agirre-etal-2015-semeval}, \citeyear{agirre-etal-2016-semeval}), \textbf{STS Benchmark}\cite{cer-etal-2017-semeval} and \textbf{SICK-Relatedness}~\cite{MARELLI14.363} using the SentEval toolkit\cite{conneau2018senteval}, following previous research. Spearman correlation based on cosine similarity is reported as the main metric~\cite{reimers-etal-2016-task}.

\textbf{Reranking \& Retrieval tasks.} For reranking tasks, the model generates sentence representations for a given query and a set of reference sentences (relevant and irrelevant), and ranks the references based on their similarity to the query representation. Retrieval tasks, similar to reranking tasks, involve the model embedding a query and documents in a corpus, and ranking the documents by similarity.  We evaluate DenoSent on 4 reranking tasks:
\textbf{AskUbuntuDupQuestions}~\cite{https://doi.org/10.48550/arxiv.1512.05726},
\textbf{MindSmallReranking}~\cite{wu-etal-2020-mind},
\textbf{SciDocsRR}~\cite{https://doi.org/10.48550/arxiv.2004.07180},
and \textbf{StackOverflowDupQuestions}~\cite{10.1145/3283812.3283815},
and a retrieval task: \textbf{QuoraRetrieval}~\cite{thakur2021beir}. We report the mean MRR@1 and MAP@1 as the main results.

\textbf{Classification tasks.} For classification tasks, each sentence in the datasets has a corresponding label. Sentence representations are obtained by the provided model and an extra logistic regression classifier is trained on these representations and their corresponding label. We evaluate DenoSent on 10 classification tasks: 
\textbf{AmazonCounterfactual}~\cite{https://doi.org/10.48550/arxiv.2104.06893}, 
\textbf{AmazonReviews}~\cite{10.1145/2507157.2507163}, 
\textbf{Banking77}~\cite{https://doi.org/10.48550/arxiv.2003.04807},
\textbf{Emotion}~\cite{saravia-etal-2018-carer},
\textbf{MassiveIntent}~\cite{https://doi.org/10.48550/arxiv.2204.08582},
\textbf{MassiveScenario}~\cite{https://doi.org/10.48550/arxiv.2204.08582},
\textbf{MTOPDomain}~\cite{https://doi.org/10.48550/arxiv.2008.09335}, \textbf{MTOPIntent}~\cite{https://doi.org/10.48550/arxiv.2008.09335},
\textbf{ToxicConversations}~\cite{kaggle_toxic},
\textbf{TweetSentimentExtraction}~\cite{kaggle_tweet}. We report the classification accuracy as the main metric.

\subsection{Baselines}
In this study, the proposed method was evaluated and compared to the following established methods.

\textbf{Glove}~\cite{pennington-etal-2014-glove} takes the Glove embedding of each word in the sentence as the sentence's representation.
\textbf{InferSent}~\cite{conneau2017supervised} uses Glove with some signal enhancement and is trained further on the NLI dataset.
\textbf{Universal Sentence Encoder}~\cite{cer-etal-2018-universal} uses the Transformer model and learns the objective of reconstructing surrounding sentences in a paragraph.
\textbf{BERT(CLS, Mean, First-Last Avg.)}~\cite{devlin2018bert} directly utilizes BERT's outputs as sentence representations, using different pooling strategies.
\textbf{BERT-Flow}~\cite{li2020sentence} reversibly maps the BERT output space from a cone to the standard Gaussian distribution space.
\textbf{BERT-Whitening}~\cite{su2021whitening} improves the quality of sentence representation by simple vector whitening.
\textbf{ConSERT}~\cite{yan2021consert} and \textbf{SimCSE}\cite{gao2021simcse} is based on contrastive learning and uses different data augmentation strategies to construct positive sentence pairs.
\textbf{DCLR}~\cite{zhou2022debiased} uses an instance weighting strategy to alleviate the false-negative problem in contrastive learning.
\textbf{DiffCSE}~\cite{chuang2022diffcse} is optimized on SimCSE to improve the effectiveness of the sentence representation model using forged  samples.
\textbf{PromptBERT}~\cite{jiang2022promptbert} uses prompts to generate sentence representations. 
\textbf{SNCSE}~\cite{wang2022sncse}  is a contrastive learning method based on soft negative examples.
\textbf{CMLM}~\cite{yang-etal-2021-universal} incorporates the learning of sentence representation into MLM training. 
\textbf{PaSeR}~\cite{wu2022sentence} proposed an intra-sentence objective that learns sentence representation by utilizing the encoded sentence representation to predict masked phrases in the input sentence.

\subsection{Implementation Details}

For the implementation of the proposed method, we use pre-trained \textit{bert-base-uncased} as the encoder and randomly initialized transformer layers as the decoder for all our experiments. We use the unsupervised Wiki dataset used in SimCSE~\cite{gao2021simcse} as our self-supervised training dataset. For back translation data augmentation, we use pre-trained machine translation models~\cite{tiedemann-thottingal-2020-opus} to translate the training sentences to Chinese and then translate them back to English. We use a learning rate of 5e-5 and AdamW~\cite{https://doi.org/10.48550/arxiv.1711.05101} as the optimizer. For the input sequence length, we use a value of 32. For the denoising objective,  we use \{0.8, 0.825, 0.85, 0.875, 0.9\} as the dropout rates for continuous perturbations, \{12, 14, 16\} as the number of decoder transformer layers and perform a sweep on these parameters then select the checkpoint that has the highest spearman correlation on the STS-Benchmark development set for evaluation. We use 0.825 as the dropout rate and 16 transformer layers for reported results. For the contrastive objective, we use a temperature $\tau=0.03$. For the pooling strategy, we fit every sentence with the same template "[X] means [MASK]." and use the encoder output vector of the [MASK] token as the sentence representation through all our experiments. We conduct all the experiments on a machine with 8 NVIDIA GeForce RTX 3090 GPUs.

\begin{figure}[t]
    \centering
    \includegraphics[height=6cm]{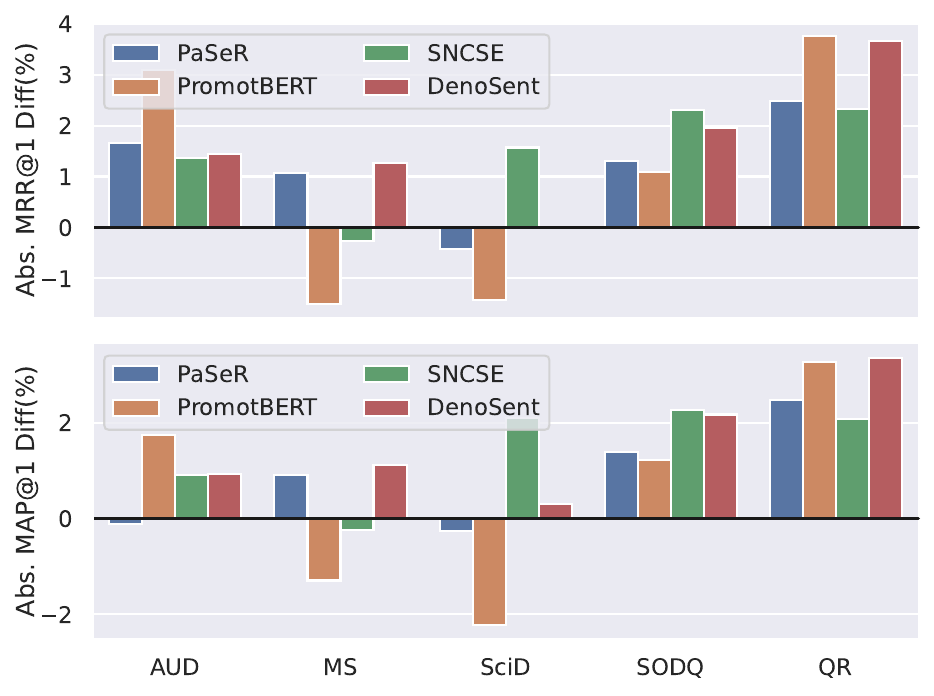}
    \caption{Absolute performance difference on reranking and retrieval tasks compared to SimCSE. AUD, MS, SciD, SODQ and QR denotes AskUbuntuDupQuestions, MindSmallReranking, SciDocsRR, StackOverflowDupQuestions and QuoraRetrieval, respectively.}
    \label{fig:reranking_results}
\end{figure}

\subsection{Main Results}

Table \ref{sts-exp} illustrates the performance of DenoSent on 7 STS tasks compared to previous methods. All experiments are conducted under a self-supervised/unsupervised setting except for non-BERT models.
The results reveal that methods that either do not use a PLM or rely solely on post-processing are less effective than those that apply contrastive and generative approaches on a PLM.
In the context of single-objective learning, the contrastive objective proves to be more effective for semantic textual similarity tasks than the generative objective since it directly optimizes representation similarities. The proposed denoising objective shows competitive performance compared to contrastive methods despite the fact that it is completely complementary to them. The utilization of the contrastive objective alone in the DenoSent model resulted in a 1.71\% absolute improvement in performance compared to the SimCSE model. This demonstrates the effectiveness of incorporating discrete noises and the [MASK] token pooling strategy, as the contrastive DenoSent model is identical to the SimCSE model in all other aspects. The proposed framework effectively integrates both inter-sentence and intra-sentence objectives, resulting in superior performance on STS tasks.

\begin{table}[h]
    \centering
    \begin{tabular}{|c|c|}
    \hline
       \textbf{Model}  &  \textbf{Avg. Classification Accuracy} \\
       \hline
        Glove & 56.42 \\
        BERT(CLS) & 60.32 \\
        SimCSE & 62.73 \\
        PaSeR & 63,23 \\
        PromptBERT & 63.78 \\
        SNCSE & 62.82 \\
        \hline
        \textbf{DenoSent} & \textbf{64.46}\\
        \hline
    \end{tabular}
    \caption{Evaluation performance on classification tasks.}
    \label{tab:classification_avg}
\end{table}

In order to assess the generalizability of DenoSent, a comprehensive set of experiments was conducted on reranking, retrieval and classification tasks. The results, as illustrated in Figure \ref{fig:reranking_results}, demonstrate that DenoSent consistently outperforms SimCSE on reranking and retrieval tasks, and exhibits a higher degree of robustness across various tasks and domains compared to other baselines. Table \ref{tab:classification_avg} presents the evaluation results for the average accuracy across 10 sentence-level classification tasks. The results indicate that DenoSent exhibits the highest performance on classification tasks, demonstrating its strong capability for generalization. The results of these tasks indicate that utilizing both intra-sentence and inter-sentence objectives not only improves performance on STS tasks, but also leads to enhancements in the overall generalizability.

\begin{table}[h]
    \centering
    \begin{tabular}{|c|c|}
    \hline
        \textbf{Model} & \textbf{Avg. STS}\\
        \hline
        \textbf{DenoSent} & \textbf{79.33} \\
        DenoSent$^{\diamondsuit}$ & 78.98 \\
        \hline
        w/o \textit{denoising} & 77.96 \\
        w/o \textit{contrastive} & 76.19 \\
        w/o \textit{discrete noise} & 77.99 \\
        w/o \textit{denoising +discrete noise} & 76.17\\
        w/o \textit{contrastive +discrete noise} & 74.54 \\
        w/ \textit{CLS Pooling} & 78.66 \\
    \hline
    \end{tabular}
    \caption{Ablation on the components in DenoSent. $\diamondsuit$ denotes using an LLM to introduce discrete noise.}
    \label{tab:ablation_table}
\end{table}

\section{Ablation Study}

\textbf{Effects of proposed components.} In Table \ref{tab:ablation_table}, we investigate the impacts of different proposed components in the DenoSent framework. The utilization of both contrastive and denoising objectives has been demonstrated to be crucial for achieving high performance. The combination of these objectives results in a significant improvement in performance. Additionally, the incorporation of discrete noises has been found to enhance performance for both objectives consistently. The utilization of [MASK] token pooling, instead of [CLS] pooling, has also been shown to provide a slight boost in performance, as previously reported in \citealp{jiang2022promptbert}.

\begin{figure}[t]
    \centering
    \begin{subfigure}[b]{0.23\textwidth}
        \centering
        \includegraphics[width=\textwidth]{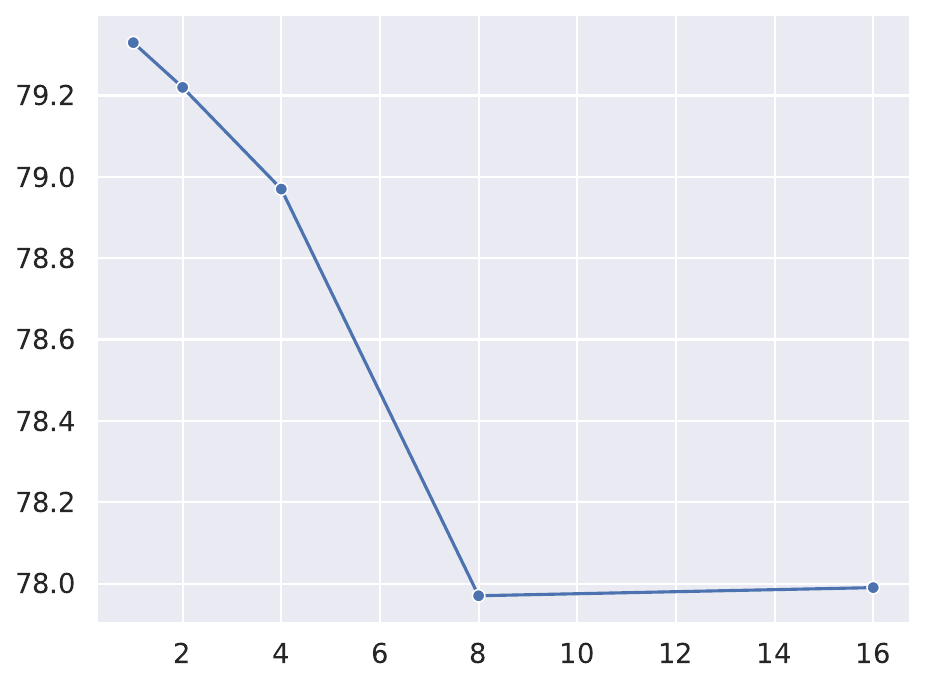}
        \caption{Impact of attention heads.}
        \label{fig:attention_head_ablation}
    \end{subfigure}
    \begin{subfigure}[b]{0.23\textwidth}
        \centering
        \includegraphics[width=\textwidth]{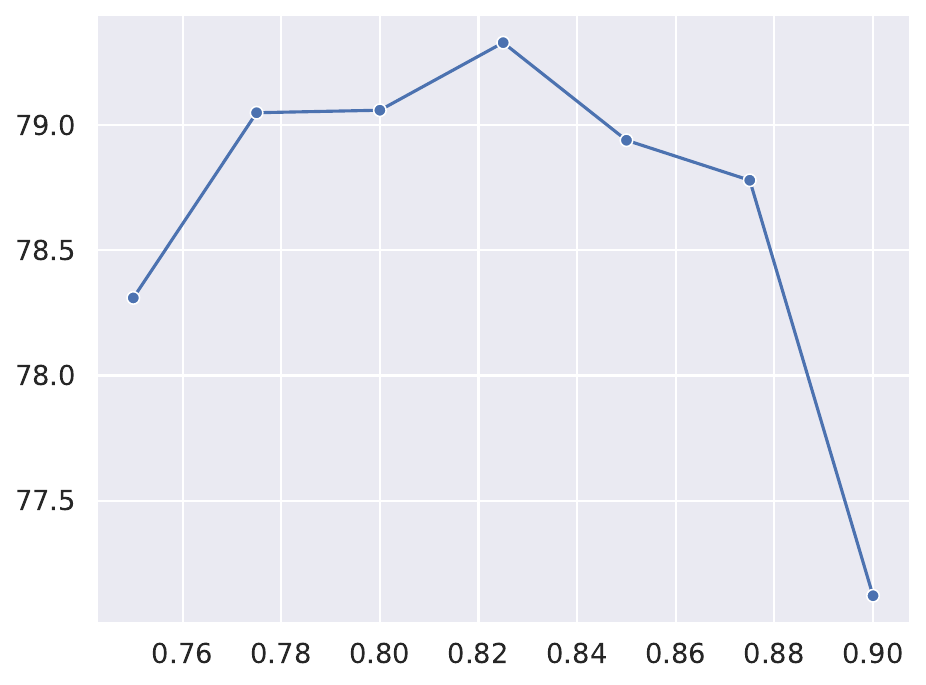}
        \caption{Impact of dropout rates.}
        \label{fig:dropout_rate_ablation}
    \end{subfigure}
    \label{fig:ablation2}
    \caption{Average STS performance using different numbers of attention heads and dropout rates.}
\end{figure}

\textbf{Effects of different number of attention heads in the decoder.} For the denoising objective, we use single-head attention instead multi-attention in our experiments. The results, depicted in Figure \ref{fig:attention_head_ablation}, indicate that an increase in the number of attention heads results in a decrease in performance.  This may be due to the fact that the multi-head attention technique enhances transformer models by offering multiple perspectives on attention. However, in the case of the DenoSent decoder, the memory input sequence length for the transformer layers is fixed at 1, rendering the utilization of multiple attention heads redundant. On the other hand, an increasing number of attention heads results in a reduction of the dimensionality of the sentence representation during computation. This decrease in dimensionality impairs the representation capabilities of the model, thereby leading to a decline in performance.

\textbf{Effects of the continuous noise level.} In the proposed method DenoSent, we employ dropout as a technique for introducing controlled corruption to sentences in the continuous space. The dropout rate is used to define the level of corruption added to the sentence. It is crucial that the injected noise is substantial enough to render the learning task sufficiently challenging, thus enabling our model to learn meaningful semantic information in sentence representations. As illustrated in Figure \ref{fig:dropout_rate_ablation}, the performance of the model is sensitive to the choice of dropout rate, with optimal results observed for moderate values. If the value is set too high, the input becomes excessively corrupted, rendering the task overly challenging and impeding the model's learning capability. Conversely, if the level of corruption is too low, the denoising task becomes overly simplistic, preventing the model from effectively leveraging the semantic information embedded in the sentence representation.

\section{Conclusion}

In this work, we introduce DenoSent, a self-supervised sentence representation learning framework that incorporates both intra-sentence and inter-sentence objectives. We propose a novel denoising objective that uses sentence representation to recover a noisy sentence input to its original. We introduce both discrete and continuous noises to perturb the input sentence to facilitate our denoising objective. Furthermore, we combine the denoising objective with the contrastive objective, allowing representations to benefit from both intra-sentence and inter-sentence supervision. We evaluate our model on numerous tasks ranging from semantic textual similarity, reranking, retrieval and classification, showing superior performance and generalization ability.

\section*{Acknowledgments}
This work was supported by the National Natural Science Foundation of China (No. 62236004 and No. 62022027). The authors would like to thank the anonymous reviewers for their comprehensive and insightful reviews and suggestions.

% \bibliography{aaai24}
% \bibliographystyle{abbrv}
{\small
	\bibliography{references}

\begin{thebibliography}{70}
\providecommand{\natexlab}[1]{#1}

\bibitem[{Agirre et~al.(2015)Agirre, Banea, Cardie, Cer, Diab, Gonzalez-Agirre, Guo, Lopez-Gazpio, Maritxalar, Mihalcea, Rigau, Uria, and Wiebe}]{agirre-etal-2015-semeval}
Agirre, E.; Banea, C.; Cardie, C.; Cer, D.; Diab, M.; Gonzalez-Agirre, A.; Guo, W.; Lopez-Gazpio, I.; Maritxalar, M.; Mihalcea, R.; Rigau, G.; Uria, L.; and Wiebe, J. 2015.
\newblock {S}em{E}val-2015 Task 2: Semantic Textual Similarity, {E}nglish, {S}panish and Pilot on Interpretability.

\bibitem[{Agirre et~al.(2014)Agirre, Banea, Cardie, Cer, Diab, Gonzalez-Agirre, Guo, Mihalcea, Rigau, and Wiebe}]{agirre-etal-2014-semeval}
Agirre, E.; Banea, C.; Cardie, C.; Cer, D.; Diab, M.; Gonzalez-Agirre, A.; Guo, W.; Mihalcea, R.; Rigau, G.; and Wiebe, J. 2014.
\newblock {S}em{E}val-2014 Task 10: Multilingual Semantic Textual Similarity.

\bibitem[{Agirre et~al.(2016)Agirre, Banea, Cer, Diab, Gonzalez-Agirre, Mihalcea, Rigau, and Wiebe}]{agirre-etal-2016-semeval}
Agirre, E.; Banea, C.; Cer, D.; Diab, M.; Gonzalez-Agirre, A.; Mihalcea, R.; Rigau, G.; and Wiebe, J. 2016.
\newblock {S}em{E}val-2016 Task 1: Semantic Textual Similarity, Monolingual and Cross-Lingual Evaluation.

\bibitem[{Agirre et~al.(2012)Agirre, Cer, Diab, and Gonzalez-Agirre}]{agirre-etal-2012-semeval}
Agirre, E.; Cer, D.; Diab, M.; and Gonzalez-Agirre, A. 2012.
\newblock {S}em{E}val-2012 Task 6: A Pilot on Semantic Textual Similarity.

\bibitem[{Agirre et~al.(2013)Agirre, Cer, Diab, Gonzalez-Agirre, and Guo}]{agirre-etal-2013-sem}
Agirre, E.; Cer, D.; Diab, M.; Gonzalez-Agirre, A.; and Guo, W. 2013.
\newblock *{SEM} 2013 shared task: Semantic Textual Similarity.

\bibitem[{Balestriero et~al.(2023)Balestriero, Ibrahim, Sobal, Morcos, Shekhar, Goldstein, Bordes, Bardes, Mialon, Tian, Schwarzschild, Wilson, Geiping, Garrido, Fernandez, Bar, Pirsiavash, LeCun, and Goldblum}]{balestriero2023cookbook}
Balestriero, R.; Ibrahim, M.; Sobal, V.; Morcos, A.; Shekhar, S.; Goldstein, T.; Bordes, F.; Bardes, A.; Mialon, G.; Tian, Y.; Schwarzschild, A.; Wilson, A.~G.; Geiping, J.; Garrido, Q.; Fernandez, P.; Bar, A.; Pirsiavash, H.; LeCun, Y.; and Goldblum, M. 2023.
\newblock A Cookbook of Self-Supervised Learning.
\newblock arXiv:2304.12210.

\bibitem[{Casanueva et~al.(2020)Casanueva, Temčinas, Gerz, Henderson, and Vulić}]{https://doi.org/10.48550/arxiv.2003.04807}
Casanueva, I.; Temčinas, T.; Gerz, D.; Henderson, M.; and Vulić, I. 2020.
\newblock Efficient Intent Detection with Dual Sentence Encoders.

\bibitem[{Cer et~al.(2017)Cer, Diab, Agirre, Lopez-Gazpio, and Specia}]{cer-etal-2017-semeval}
Cer, D.; Diab, M.; Agirre, E.; Lopez-Gazpio, I.; and Specia, L. 2017.
\newblock {S}em{E}val-2017 Task 1: Semantic Textual Similarity Multilingual and Crosslingual Focused Evaluation.

\bibitem[{Cer et~al.(2018)Cer, Yang, Kong, Hua, Limtiaco, St.~John, Constant, Guajardo-Cespedes, Yuan, Tar, Strope, and Kurzweil}]{cer-etal-2018-universal}
Cer, D.; Yang, Y.; Kong, S.-y.; Hua, N.; Limtiaco, N.; St.~John, R.; Constant, N.; Guajardo-Cespedes, M.; Yuan, S.; Tar, C.; Strope, B.; and Kurzweil, R. 2018.
\newblock Universal Sentence Encoder for {E}nglish.
\newblock In \emph{Proceedings of the 2018 Conference on Empirical Methods in Natural Language Processing: System Demonstrations}.

\bibitem[{Chen et~al.(2020{\natexlab{a}})Chen, Kornblith, Norouzi, and Hinton}]{chen2020simple}
Chen, T.; Kornblith, S.; Norouzi, M.; and Hinton, G. 2020{\natexlab{a}}.
\newblock A simple framework for contrastive learning of visual representations.
\newblock In \emph{International conference on machine learning}.

\bibitem[{Chen et~al.(2020{\natexlab{b}})Chen, Kornblith, Norouzi, and Hinton}]{https://doi.org/10.48550/arxiv.2002.05709}
Chen, T.; Kornblith, S.; Norouzi, M.; and Hinton, G. 2020{\natexlab{b}}.
\newblock A Simple Framework for Contrastive Learning of Visual Representations.

\bibitem[{Cheng et~al.(2023)Cheng, Yang, Sun, Li, and Qiu}]{cheng2023improving}
Cheng, Q.; Yang, X.; Sun, T.; Li, L.; and Qiu, X. 2023.
\newblock Improving Contrastive Learning of Sentence Embeddings from AI Feedback.
\newblock arXiv:2305.01918.

\bibitem[{Chuang et~al.(2022)Chuang, Dangovski, Luo, Zhang, Chang, Soljačić, Li, tau Yih, Kim, and Glass}]{chuang2022diffcse}
Chuang, Y.-S.; Dangovski, R.; Luo, H.; Zhang, Y.; Chang, S.; Soljačić, M.; Li, S.-W.; tau Yih, W.; Kim, Y.; and Glass, J. 2022.
\newblock DiffCSE: Difference-based Contrastive Learning for Sentence Embeddings.
\newblock arXiv:2204.10298.

\bibitem[{Cohan et~al.(2020)Cohan, Feldman, Beltagy, Downey, and Weld}]{https://doi.org/10.48550/arxiv.2004.07180}
Cohan, A.; Feldman, S.; Beltagy, I.; Downey, D.; and Weld, D.~S. 2020.
\newblock SPECTER: Document-level Representation Learning using Citation-informed Transformers.

\bibitem[{Conneau and Kiela(2018)}]{conneau2018senteval}
Conneau, A.; and Kiela, D. 2018.
\newblock SentEval: An Evaluation Toolkit for Universal Sentence Representations.
\newblock \emph{arXiv preprint arXiv:1803.05449}.

\bibitem[{Conneau et~al.(2017)Conneau, Kiela, Schwenk, Barrault, and Bordes}]{conneau2017supervised}
Conneau, A.; Kiela, D.; Schwenk, H.; Barrault, L.; and Bordes, A. 2017.
\newblock Supervised Learning of Universal Sentence Representations from Natural Language Inference Data.
\newblock arXiv:1705.02364.

\bibitem[{Devlin et~al.(2018)Devlin, Chang, Lee, and Toutanova}]{devlin2018bert}
Devlin, J.; Chang, M.-W.; Lee, K.; and Toutanova, K. 2018.
\newblock Bert: Pre-training of deep bidirectional transformers for language understanding.
\newblock \emph{arXiv preprint arXiv:1810.04805}.

\bibitem[{Ethayarajh(2019)}]{ethayarajh2019contextual}
Ethayarajh, K. 2019.
\newblock How contextual are contextualized word representations? comparing the geometry of BERT, ELMo, and GPT-2 embeddings.
\newblock \emph{arXiv preprint arXiv:1909.00512}.

\bibitem[{FitzGerald et~al.(2022)FitzGerald, Hench, Peris, Mackie, Rottmann, Sanchez, Nash, Urbach, Kakarala, Singh, Ranganath, Crist, Britan, Leeuwis, Tur, and Natarajan}]{https://doi.org/10.48550/arxiv.2204.08582}
FitzGerald, J.; Hench, C.; Peris, C.; Mackie, S.; Rottmann, K.; Sanchez, A.; Nash, A.; Urbach, L.; Kakarala, V.; Singh, R.; Ranganath, S.; Crist, L.; Britan, M.; Leeuwis, W.; Tur, G.; and Natarajan, P. 2022.
\newblock MASSIVE: A 1M-Example Multilingual Natural Language Understanding Dataset with 51 Typologically-Diverse Languages.

\bibitem[{Gao, Yao, and Chen(2021)}]{gao2021simcse}
Gao, T.; Yao, X.; and Chen, D. 2021.
\newblock Simcse: Simple contrastive learning of sentence embeddings.
\newblock \emph{arXiv preprint arXiv:2104.08821}.

\bibitem[{Giorgi et~al.(2021)Giorgi, Nitski, Wang, and Bader}]{2021DeCLUTR}
Giorgi, J.; Nitski, O.; Wang, B.; and Bader, G. 2021.
\newblock DeCLUTR: Deep Contrastive Learning for Unsupervised Textual Representations.
\newblock In \emph{Meeting of the Association for Computational Linguistics}.

\bibitem[{Hill, Cho, and Korhonen(2016)}]{hill2016learning}
Hill, F.; Cho, K.; and Korhonen, A. 2016.
\newblock Learning distributed representations of sentences from unlabelled data.
\newblock \emph{arXiv preprint arXiv:1602.03483}.

\bibitem[{Janson et~al.(2021)Janson, Gogoulou, Ylip{\"a}{\"a}, Cuba~Gyllensten, and Sahlgren}]{janson2021semantic}
Janson, S.; Gogoulou, E.; Ylip{\"a}{\"a}, E.; Cuba~Gyllensten, A.; and Sahlgren, M. 2021.
\newblock Semantic re-tuning with contrastive tension.
\newblock In \emph{International Conference on Learning Representations, 2021}.

\bibitem[{Jiang et~al.(2022)Jiang, Huang, Zhang, Wang, Zhuang, Wei, Huang, Zhang, and Zhang}]{jiang2022promptbert}
Jiang, T.; Huang, S.; Zhang, Z.; Wang, D.; Zhuang, F.; Wei, F.; Huang, H.; Zhang, L.; and Zhang, Q. 2022.
\newblock PromptBERT: Improving BERT Sentence Embeddings with Prompts.
\newblock \emph{arXiv preprint arXiv:2201.04337}.

\bibitem[{Kaggle(2019)}]{kaggle_toxic}
Kaggle. 2019.
\newblock ToxicConversations.

\bibitem[{Kaggle(2020)}]{kaggle_tweet}
Kaggle. 2020.
\newblock TweetSentimentExtraction.

\bibitem[{Kim, Yoo, and Lee(2021)}]{kim-etal-2021-self}
Kim, T.; Yoo, K.~M.; and Lee, S.-g. 2021.
\newblock Self-Guided Contrastive Learning for {BERT} Sentence Representations.
\newblock In \emph{Proceedings of the 59th Annual Meeting of the Association for Computational Linguistics and the 11th International Joint Conference on Natural Language Processing (Volume 1: Long Papers)}.

\bibitem[{Kingma and Welling(2013)}]{kingma2013auto}
Kingma, D.~P.; and Welling, M. 2013.
\newblock Auto-encoding variational bayes.
\newblock \emph{arXiv preprint arXiv:1312.6114}.

\bibitem[{Kiros et~al.(2015)Kiros, Zhu, Salakhutdinov, Zemel, Urtasun, Torralba, and Fidler}]{kiros2015skip}
Kiros, R.; Zhu, Y.; Salakhutdinov, R.~R.; Zemel, R.; Urtasun, R.; Torralba, A.; and Fidler, S. 2015.
\newblock Skip-thought vectors.
\newblock \emph{Advances in neural information processing systems}.

\bibitem[{Lei et~al.(2015)Lei, Joshi, Barzilay, Jaakkola, Tymoshenko, Moschitti, and Marquez}]{https://doi.org/10.48550/arxiv.1512.05726}
Lei, T.; Joshi, H.; Barzilay, R.; Jaakkola, T.; Tymoshenko, K.; Moschitti, A.; and Marquez, L. 2015.
\newblock Semi-supervised Question Retrieval with Gated Convolutions.

\bibitem[{Li et~al.(2020{\natexlab{a}})Li, Zhou, He, Wang, Yang, and Li}]{li2020sentence}
Li, B.; Zhou, H.; He, J.; Wang, M.; Yang, Y.; and Li, L. 2020{\natexlab{a}}.
\newblock On the Sentence Embeddings from Pre-trained Language Models.
\newblock arXiv:2011.05864.

\bibitem[{Li et~al.(2020{\natexlab{b}})Li, Arora, Chen, Gupta, Gupta, and Mehdad}]{https://doi.org/10.48550/arxiv.2008.09335}
Li, H.; Arora, A.; Chen, S.; Gupta, A.; Gupta, S.; and Mehdad, Y. 2020{\natexlab{b}}.
\newblock MTOP: A Comprehensive Multilingual Task-Oriented Semantic Parsing Benchmark.

\bibitem[{Liu et~al.(2018)Liu, Wang, Leng, and Zhai}]{10.1145/3283812.3283815}
Liu, X.; Wang, C.; Leng, Y.; and Zhai, C. 2018.
\newblock LinkSO: A Dataset for Learning to Retrieve Similar Question Answer Pairs on Software Development Forums.
\newblock In \emph{Proceedings of the 4th ACM SIGSOFT International Workshop on NLP for Software Engineering}.

\bibitem[{Liu et~al.(2021)Liu, Zhang, Hou, Mian, Wang, Zhang, and Tang}]{Liu_2021}
Liu, X.; Zhang, F.; Hou, Z.; Mian, L.; Wang, Z.; Zhang, J.; and Tang, J. 2021.
\newblock Self-supervised Learning: Generative or Contrastive.
\newblock \emph{{IEEE} Transactions on Knowledge and Data Engineering}.

\bibitem[{Liu et~al.(2019)Liu, Ott, Goyal, Du, Joshi, Chen, Levy, Lewis, Zettlemoyer, and Stoyanov}]{liu2019roberta}
Liu, Y.; Ott, M.; Goyal, N.; Du, J.; Joshi, M.; Chen, D.; Levy, O.; Lewis, M.; Zettlemoyer, L.; and Stoyanov, V. 2019.
\newblock Roberta: A robustly optimized bert pretraining approach.
\newblock \emph{arXiv preprint arXiv:1907.11692}.

\bibitem[{Logeswaran and Lee(2018)}]{logeswaran2018efficient}
Logeswaran, L.; and Lee, H. 2018.
\newblock An efficient framework for learning sentence representations.
\newblock \emph{arXiv preprint arXiv:1803.02893}.

\bibitem[{Loshchilov and Hutter(2017)}]{https://doi.org/10.48550/arxiv.1711.05101}
Loshchilov, I.; and Hutter, F. 2017.
\newblock Decoupled Weight Decay Regularization.

\bibitem[{Marelli et~al.(2014)Marelli, Menini, Baroni, Bentivogli, Bernardi, and Zamparelli}]{MARELLI14.363}
Marelli, M.; Menini, S.; Baroni, M.; Bentivogli, L.; Bernardi, R.; and Zamparelli, R. 2014.
\newblock A SICK Cure for the Evaluation of Compositional Distributional Semantic Models.
\newblock In \emph{Proceedings of the Ninth International Conference on Language Resources and Evaluation (LREC'14)}.

\bibitem[{McAuley and Leskovec(2013)}]{10.1145/2507157.2507163}
McAuley, J.; and Leskovec, J. 2013.
\newblock Hidden Factors and Hidden Topics: Understanding Rating Dimensions with Review Text.
\newblock In \emph{Proceedings of the 7th ACM Conference on Recommender Systems}.

\bibitem[{Meng et~al.(2021)Meng, Xiong, Bajaj, Tiwary, Bennett, Han, and Song}]{meng2021cocolm}
Meng, Y.; Xiong, C.; Bajaj, P.; Tiwary, S.; Bennett, P.; Han, J.; and Song, X. 2021.
\newblock COCO-LM: Correcting and Contrasting Text Sequences for Language Model Pretraining.
\newblock arXiv:2102.08473.

\bibitem[{Mikolov et~al.(2013)Mikolov, Chen, Corrado, and Dean}]{mikolov2013efficient}
Mikolov, T.; Chen, K.; Corrado, G.; and Dean, J. 2013.
\newblock Efficient estimation of word representations in vector space.
\newblock \emph{arXiv preprint arXiv:1301.3781}.

\bibitem[{Muennighoff et~al.(2022)Muennighoff, Tazi, Magne, and Reimers}]{muennighoff2022mteb}
Muennighoff, N.; Tazi, N.; Magne, L.; and Reimers, N. 2022.
\newblock MTEB: Massive Text Embedding Benchmark.
\newblock \emph{arXiv preprint arXiv:2210.07316}.

\bibitem[{O'Neill et~al.(2021)O'Neill, Rozenshtein, Kiryo, Kubota, and Bollegala}]{https://doi.org/10.48550/arxiv.2104.06893}
O'Neill, J.; Rozenshtein, P.; Kiryo, R.; Kubota, M.; and Bollegala, D. 2021.
\newblock I Wish I Would Have Loved This One, But I Didn't -- A Multilingual Dataset for Counterfactual Detection in Product Reviews.

\bibitem[{Oord, Li, and Vinyals(2018)}]{https://doi.org/10.48550/arxiv.1807.03748}
Oord, A. v.~d.; Li, Y.; and Vinyals, O. 2018.
\newblock Representation Learning with Contrastive Predictive Coding.

\bibitem[{Ouyang et~al.(2022)Ouyang, Wu, Jiang, Almeida, Wainwright, Mishkin, Zhang, Agarwal, Slama, Ray, Schulman, Hilton, Kelton, Miller, Simens, Askell, Welinder, Christiano, Leike, and Lowe}]{ouyang2022training}
Ouyang, L.; Wu, J.; Jiang, X.; Almeida, D.; Wainwright, C.~L.; Mishkin, P.; Zhang, C.; Agarwal, S.; Slama, K.; Ray, A.; Schulman, J.; Hilton, J.; Kelton, F.; Miller, L.; Simens, M.; Askell, A.; Welinder, P.; Christiano, P.; Leike, J.; and Lowe, R. 2022.
\newblock Training language models to follow instructions with human feedback.
\newblock arXiv:2203.02155.

\bibitem[{Pennington, Socher, and Manning(2014)}]{pennington-etal-2014-glove}
Pennington, J.; Socher, R.; and Manning, C. 2014.
\newblock {G}lo{V}e: Global Vectors for Word Representation.
\newblock In \emph{Proceedings of the 2014 Conference on Empirical Methods in Natural Language Processing ({EMNLP})}.

\bibitem[{Reimers, Beyer, and Gurevych(2016)}]{reimers-etal-2016-task}
Reimers, N.; Beyer, P.; and Gurevych, I. 2016.
\newblock Task-Oriented Intrinsic Evaluation of Semantic Textual Similarity.
\newblock In \emph{Proceedings of {COLING} 2016, the 26th International Conference on Computational Linguistics: Technical Papers}.

\bibitem[{Reimers and Gurevych(2019)}]{2019Sentence}
Reimers, N.; and Gurevych, I. 2019.
\newblock Sentence-BERT: Sentence Embeddings using Siamese BERT-Networks.

\bibitem[{Saravia et~al.(2018)Saravia, Liu, Huang, Wu, and Chen}]{saravia-etal-2018-carer}
Saravia, E.; Liu, H.-C.~T.; Huang, Y.-H.; Wu, J.; and Chen, Y.-S. 2018.
\newblock {CARER}: Contextualized Affect Representations for Emotion Recognition.
\newblock In \emph{Proceedings of the 2018 Conference on Empirical Methods in Natural Language Processing}.

\bibitem[{Schneider et~al.(2019)Schneider, Baevski, Collobert, and Auli}]{schneider2019wav2vec}
Schneider, S.; Baevski, A.; Collobert, R.; and Auli, M. 2019.
\newblock wav2vec: Unsupervised Pre-training for Speech Recognition.
\newblock arXiv:1904.05862.

\bibitem[{Srivastava et~al.(2014)Srivastava, Hinton, Krizhevsky, Sutskever, and Salakhutdinov}]{srivastava2014dropout}
Srivastava, N.; Hinton, G.~E.; Krizhevsky, A.; Sutskever, I.; and Salakhutdinov, R. 2014.
\newblock Dropout: a simple way to prevent neural networks from overfitting.
\newblock \emph{Journal of Machine Learning Research}.

\bibitem[{Su et~al.(2022)Su, Kasai, Wang, Hu, Ostendorf, Yih, Smith, Zettlemoyer, Yu et~al.}]{su2022one}
Su, H.; Kasai, J.; Wang, Y.; Hu, Y.; Ostendorf, M.; Yih, W.-t.; Smith, N.~A.; Zettlemoyer, L.; Yu, T.; et~al. 2022.
\newblock One embedder, any task: Instruction-finetuned text embeddings.
\newblock \emph{arXiv preprint arXiv:2212.09741}.

\bibitem[{Su et~al.(2021)Su, Cao, Liu, and Ou}]{su2021whitening}
Su, J.; Cao, J.; Liu, W.; and Ou, Y. 2021.
\newblock Whitening Sentence Representations for Better Semantics and Faster Retrieval.
\newblock arXiv:2103.15316.

\bibitem[{Sung et~al.(2018)Sung, Yang, Zhang, Xiang, Torr, and Hospedales}]{sung2018learning}
Sung, F.; Yang, Y.; Zhang, L.; Xiang, T.; Torr, P.~H.; and Hospedales, T.~M. 2018.
\newblock Learning to compare: Relation network for few-shot learning.
\newblock In \emph{Proceedings of the IEEE conference on computer vision and pattern recognition}.

\bibitem[{Thakur et~al.(2021)Thakur, Reimers, R{\"u}ckl{\'e}, Srivastava, and Gurevych}]{thakur2021beir}
Thakur, N.; Reimers, N.; R{\"u}ckl{\'e}, A.; Srivastava, A.; and Gurevych, I. 2021.
\newblock {BEIR}: A Heterogeneous Benchmark for Zero-shot Evaluation of Information Retrieval Models.
\newblock In \emph{Thirty-fifth Conference on Neural Information Processing Systems Datasets and Benchmarks Track (Round 2)}.

\bibitem[{Tiedemann and Thottingal(2020)}]{tiedemann-thottingal-2020-opus}
Tiedemann, J.; and Thottingal, S. 2020.
\newblock {OPUS}-{MT} {--} Building open translation services for the World.
\newblock In \emph{Proceedings of the 22nd Annual Conference of the European Association for Machine Translation}.

\bibitem[{Vaswani et~al.(2017)Vaswani, Shazeer, Parmar, Uszkoreit, Jones, Gomez, Kaiser, and Polosukhin}]{vaswani2017attention}
Vaswani, A.; Shazeer, N.; Parmar, N.; Uszkoreit, J.; Jones, L.; Gomez, A.~N.; Kaiser, {\L}.; and Polosukhin, I. 2017.
\newblock Attention is all you need.
\newblock \emph{Advances in neural information processing systems}, 30.

\bibitem[{Wang et~al.(2022{\natexlab{a}})Wang, Li, Huang, Dou, Kong, and Shao}]{wang2022sncse}
Wang, H.; Li, Y.; Huang, Z.; Dou, Y.; Kong, L.; and Shao, J. 2022{\natexlab{a}}.
\newblock SNCSE: Contrastive Learning for Unsupervised Sentence Embedding with Soft Negative Samples.
\newblock arXiv:2201.05979.

\bibitem[{Wang, Reimers, and Gurevych(2021)}]{wang2021tsdae}
Wang, K.; Reimers, N.; and Gurevych, I. 2021.
\newblock TSDAE: Using Transformer-based Sequential Denoising Auto-Encoder for Unsupervised Sentence Embedding Learning.
\newblock arXiv:2104.06979.

\bibitem[{Wang et~al.(2022{\natexlab{b}})Wang, Yang, Huang, Jiao, Yang, Jiang, Majumder, and Wei}]{wang2022text}
Wang, L.; Yang, N.; Huang, X.; Jiao, B.; Yang, L.; Jiang, D.; Majumder, R.; and Wei, F. 2022{\natexlab{b}}.
\newblock Text embeddings by weakly-supervised contrastive pre-training.
\newblock \emph{arXiv preprint arXiv:2212.03533}.

\bibitem[{Wang and Isola(2020)}]{wang2020understanding}
Wang, T.; and Isola, P. 2020.
\newblock Understanding contrastive representation learning through alignment and uniformity on the hypersphere.
\newblock In \emph{International Conference on Machine Learning}.

\bibitem[{Wei et~al.(2022)Wei, Bosma, Zhao, Guu, Yu, Lester, Du, Dai, and Le}]{wei2022finetuned}
Wei, J.; Bosma, M.; Zhao, V.~Y.; Guu, K.; Yu, A.~W.; Lester, B.; Du, N.; Dai, A.~M.; and Le, Q.~V. 2022.
\newblock Finetuned Language Models Are Zero-Shot Learners.
\newblock arXiv:2109.01652.

\bibitem[{Wu and Zhao(2022)}]{wu2022sentence}
Wu, B.; and Zhao, H. 2022.
\newblock Sentence Representation Learning with Generative Objective rather than Contrastive Objective.
\newblock arXiv:2210.08474.

\bibitem[{Wu et~al.(2020)Wu, Qiao, Chen, Wu, Qi, Lian, Liu, Xie, Gao, Wu, and Zhou}]{wu-etal-2020-mind}
Wu, F.; Qiao, Y.; Chen, J.-H.; Wu, C.; Qi, T.; Lian, J.; Liu, D.; Xie, X.; Gao, J.; Wu, W.; and Zhou, M. 2020.
\newblock {MIND}: A Large-scale Dataset for News Recommendation.
\newblock In \emph{Proceedings of the 58th Annual Meeting of the Association for Computational Linguistics}.

\bibitem[{Yan et~al.(2021)Yan, Li, Wang, Zhang, Wu, and Xu}]{yan2021consert}
Yan, Y.; Li, R.; Wang, S.; Zhang, F.; Wu, W.; and Xu, W. 2021.
\newblock Consert: A contrastive framework for self-supervised sentence representation transfer.
\newblock \emph{arXiv preprint arXiv:2105.11741}.

\bibitem[{Yang et~al.(2020)Yang, Yang, Cer, Law, and Darve}]{yang2020universal}
Yang, Z.; Yang, Y.; Cer, D.; Law, J.; and Darve, E. 2020.
\newblock Universal sentence representation learning with conditional masked language model.
\newblock \emph{arXiv preprint arXiv:2012.14388}.

\bibitem[{Yang et~al.(2021)Yang, Yang, Cer, Law, and Darve}]{yang-etal-2021-universal}
Yang, Z.; Yang, Y.; Cer, D.; Law, J.; and Darve, E. 2021.
\newblock Universal Sentence Representation Learning with Conditional Masked Language Model.
\newblock In \emph{Proceedings of the 2021 Conference on Empirical Methods in Natural Language Processing}.

\bibitem[{Zhang et~al.(2020)Zhang, He, Liu, Lim, and Bing}]{zhang-etal-2020-unsupervised}
Zhang, Y.; He, R.; Liu, Z.; Lim, K.~H.; and Bing, L. 2020.
\newblock An Unsupervised Sentence Embedding Method by Mutual Information Maximization.
\newblock In \emph{Proceedings of the 2020 Conference on Empirical Methods in Natural Language Processing (EMNLP)}.

\bibitem[{Zhang et~al.(2022)Zhang, Zhu, Wang, Xu, Li, and Zhao}]{zhang2022contrastive}
Zhang, Y.; Zhu, H.; Wang, Y.; Xu, N.; Li, X.; and Zhao, B. 2022.
\newblock A Contrastive Framework for Learning Sentence Representations from Pairwise and Triple-wise Perspective in Angular Space.
\newblock In \emph{Proceedings of the 60th Annual Meeting of the Association for Computational Linguistics (Volume 1: Long Papers)}.

\bibitem[{Zhou et~al.(2022)Zhou, Zhang, Zhao, and Wen}]{zhou2022debiased}
Zhou, K.; Zhang, B.; Zhao, W.~X.; and Wen, J.-R. 2022.
\newblock Debiased Contrastive Learning of Unsupervised Sentence Representations.
\newblock \emph{arXiv preprint arXiv:2205.00656}.

\end{thebibliography}
}

\end{document}